# ADVANCES ON CONCEPT DRIFT DETECTION IN REGRESSION TASKS USING SOCIAL NETWORKS THEORY

*Jean Paul Barddal, Programa de Pós-Graduação em Informática, Pontifícia Universidade Católica do Paraná, Brasil*

*Heitor Murilo Gomes, Programa de Pós-Graduação em Informática, Pontifícia Universidade Católica do Paraná, Brasil*

*Fabrício Enembreck, Programa de Pós-Graduação em Informática, Pontifícia Universidade Católica do Paraná, Brasil*

## ABSTRACT

*Mining data streams is one of the main studies in machine learning area due to its application in many knowledge areas. One of the major challenges on mining data streams is concept drift, which requires the learner to discard the current concept and adapt to a new one. Ensemble-based drift detection algorithms have been used successfully to the classification task but usually maintain a fixed size ensemble of learners running the risk of needlessly spending processing time and memory. In this paper we present improvements to the Scale-free Network Regressor (SFNR), a dynamic ensemble-based method for regression that employs social networks theory. In order to detect concept drifts SFNR uses the Adaptive Window (ADWIN) algorithm. Results show improvements in accuracy, especially in concept drift situations and better performance compared to other state-of-the-art algorithms in both real and synthetic data.*

*Keywords:   Supervised Learning, Data Stream Mining, Machine Learning, Regression, Ensemble-based Methods, Social Network Analysis*

## INTRODUCTION

One of the main subjects in machine learning is concept learning. Usually, a learner is presented to a finite amount of labeled instances generating a hypothesis to predict labels of unlabeled instances. However, concept learning may rely on a context which is not present in the initial training dataset, e.g. prediction rules which depend on the season. Changes in the context may cause changes in the concept to be learned, phenomenon known as concept drift (Bifet, Holmes, & Pfahringer, 2010). Thus, an inductor created to learn from data streams must be able to detect these drifts, adapting its hypothesis in response to the new concept. Also, it must be robust enough to discriminate between true concept drifts and noise. To cope with concept drift, ensemble-based methods were proven a good approach (Oza & Russell, 2001; Bifet, Holmes, & Pfahringer, 2010). These methods maintain a set of experts and combine their predictions in order to obtain a global prediction. The maintenance of the ensemble (addition and removal of experts) depends on each algorithm. Yet, the majority of the ensemble-based methods keep a static number of experts. Therefore, how many experts will compose the ensemble depends on parameterization prior to execution which can lead to less or more experts than are needed for effectively and efficiently representing the given data stream, i.e. if too few experts are created perhaps the accuracy will be suboptimal, conversely if too many experts are created then memory and processing time will be negatively affected. In addition, the state-of-the-art ensemble-based algorithms are not adapted to the regression task since their heuristics were developed for classification. In our approach the ensemble is a network of experts that evolves naturally accordingly to the Scale-free model (Albert & Barabási, 2002). The network allows the usage of centrality metrics, which determine the importance of each actor in the network and are used to poll experts' votes. In addition, we propose the usage of a detector, namely ADWIN (Bifet & Gavaldà, 2007), to detect concept drifts.

The remainder of this work is organized as follows. In Section 2 we present related algorithms for data stream regression. Section 3 introduces the main aspects of social networks relevant to this work. Section 4 discusses the SFNR algorithm while Section 5 focuses on the improvements of the SFNR, namely SFNR+ADWIN. Section 6 presents the experimental evaluation and a discussion over the results obtained. Finally, Section 7 presents our conclusions and future work.

## RELATED WORK

Most of the existing works on ensembles rely on developing algorithms to improve overall accuracy coping with concept drift explicitly (Bifet, Holmes, Pfahringer, Kirkby, & Gavaldà, 2009) or implicitly (Kolter & Maloof, 2005; Widmer & Kubate, 1996). Authors in (Kuncheva, 2004) shows that an ensemble can surpass an individual expert's accuracy if its component experts are diverse. An ensemble is said to be diverse if its members misclassify different instances (in regression tasks if they predict instances with different values). Another important trait of an ensemble refers to how it combines individual decisions. If the combination strategy fails to highlight correct and obfuscate incorrect decisions then the method is jeopardized. In the remainder of this section we present the state-of-the-art algorithms for data stream regression, including single classifier and ensemble methods.

**Moving Average**

The Moving Average is one of the oldest indicators of technical analysis for stock market forecasting (Brockwell & Davis, 2002). Its computation is based on a weighted average of historic stock values. We chose the Exponential Moving Average (EMA) since a conventional Moving Average takes too long to predict market tendencies. Equation 1 presents the Exponential Moving Average computation where $p_{t-1}$ stands for the price of a given stock in a time $(t-1)$ and $w$ is the algorithm's window size.

$$EMA_t = (p_{t-1} - EMA_{t-1}) \cdot \frac{2}{w+1} + EMA_{t-1} \quad (1)$$

**FIMT-DD**

Fast and Incremental Model Trees with Drift Detection (FIMT-DD) performs in real-time, observing each instance only once and maintaining a model tree (Ikonomovska, Gama, & Džeroski, 2011). The tree leaves contain linear models induced from the instances assigned to them, a process with low complexity. The algorithm has a drift detection mechanism, which adapts the model learned by updating the tree structure, enabling it to maintain accuracy during drifts.

**Adaptive Model Rules**

Adaptive Model Rules (AMRules) learns ordered and unordered rule set from data streams (Almeida, Ferreira, & Gama, 2013). These rules use a Page-Hinkley test (Mouss, Mouss, Mouss, & Sefouhi, 2004) to detect changes in the process which generates data, thus, enables the algorithm to react to drifts by pruning and enlarging the rule set.

**IBLStreams**

In (Shaker & Hüllermeier, 2012) authors presented an induction instance-based algorithm (IBLStreams) for both classification and regression. For regression problems, this algorithm uses the Root Mean Squared Error (RMSE) to determine how many neighbors will be used for resolving an instance class value, namely AdaptK. Another important feature of IBLStreams is the usage of the windowing technique, where older instances are eliminated.

**AddExp**

AddExp bounds algorithm's performance over changing concepts relative to the actual performance of an online learner trained on each concept individually (Kolter & Maloof, 2005). Authors demonstrated that AddExp suffers a loss of $\mathcal{O}(l)$ on any concept, where $l$ is the loss of a base learner when trained only on a single concept.

## SOCIAL NETWORKS

Social Networks Theory has been applied in many research fields, mainly due to its precise and formal description of structural variables. Regardless of social network analysis to subjective topics, such as individual behavior in society, the network can be precisely represented computationally as a graph. A social network is characterized by a set of actors, one or more relations and a set of connections between pairs of actors for each relation (Albert & Barabási, 2002).

There are three major network models presented in the bibliography, namely Random, Small-world and Scale-free. The Random Networks (Erdos & Rényi, 1960), (see Figure 1-a), whose construction is based on the hypothesis that the existence of a connection between a pair of nodes is given by a global probability $p$.

Figure 1. Network models.

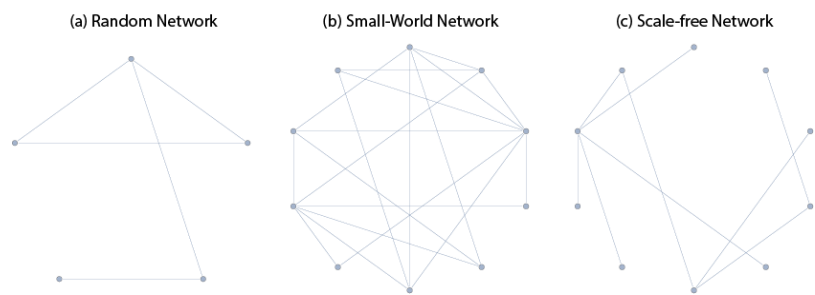

(a) Random Network  (b) Small-World Network  (c) Scale-free Network

In Small-World model (Watts & Strogatz, 1998), most actors are not neighbors of one another, yet, most of them can be reached from every other by a small number of steps (Figure 1-b) as experiments conducted in (Milgram, 1967). This model ties in characteristics of both random and regular networks. Thus, this topology presents a high clustering coefficient (as regular networks) and a small average path length (as random networks).

Finally, the Scale-free (Albert & Barabási, 2002) model aims on modeling real-world networks more accurately than Random and Small-world models. An example of Scale-free network topology is presented in Figure 1-c. Authors designed the construction (assembly) and evolution (growth) of the network as follows. The growth element defines that starting with a determined network size ($n$), for every $t$ time unit, a new actor is added to the network establishing connections with other actors. The assembly element defines the preferential attachment process. When choosing the actors which the new actor will connect to, it is assumed a probability $\Pi$ that states the chances of each actor receives a new connection.

The value of $\Pi$ depends on the actor's degree centrality and is calculated by Equation 2 where $d_r$ stands for the degree metric for the $r^{th}$ actor and $\sum_j^N d_j$ is the sum of the degree metrics for each actor in the network $N$.

$$\Pi(r) = \frac{d_r}{\sum_j^N d_j} \quad (2)$$

Due to the preferential attachment process, Scale-free networks are "dominated" by few vertices namely hubs (Correa, Crnovrsanin, & Ma, 2012). Thus, its network degree distribution follows the exponential law $p(k) \sim k^{-\lambda}$ where $p(k)$ is the probability of a random node being connected to $k$ other nodes where $\{\lambda \in \mathfrak{R}^+ \mid 2 \leq \lambda \leq 3\}$ in many real networks. Figure 2 shows an example of a Scale-free network where nodes are squares and its hubs are highlighted in darker colors.

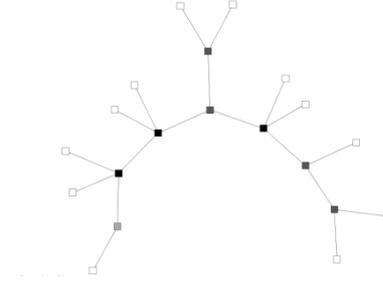

Figure 2. Scale-free network and its hubs.

In order to determine the importance of actors in social networks centrality metrics were developed. These metrics determine quantitatively the importance of actors in a social network and which are more prominent than others. Centrality metrics such as Degree, Betweenness, Closeness, Eigenvector and PageRank are discussed in (Newman, 2010).

**PRELIMINARIES**

Many ensemble-based methods presented in literature associate each expert a weight. This weight is usually used to determine the importance of an expert's vote in polling a new instance prediction. This is the case of AddExp (Kolter & Maloof, 2005). Nevertheless, the ensemble is static (does not grow and shrink) and there is no representation of the relationships between experts.

The algorithm presented in (Barddal & Enembreck, 2013) is based in an evolving network model which is able to maintain a dynamic-sized ensemble. The representation between experts is also explored and discussed in (Gomes & Enembreck, 2014; Barddal, Gomes, & Enembreck, 2014).

The Scale-free Network Regressor (SNFR) was presented at (Barddal & Enembreck, 2013) as an ensemble-based regressor which can detect and recover from concept drifts. Also, experts are represented as nodes and they establish connections following a Scale-free network model.

Whilst many experts would need many instances to detect and adapt to a concept drift, SFNR maximizes centrality metrics of hubs, thus, simple reconnections between nodes may turn an underrated node into a hub, without the need for great amounts of additions and removals in the ensemble or waiting for weight decays.

In conventional Scale-free networks, the probability of a given node establishing a new connection is proportional to its degree metric. Conversely, we adapt this probability according to the problem at hand, i.e., concept learning. Instead of a node being more prominent accordingly to its degree metric, it will be more prominent inversely proportionally to its local RMSE metric $\phi$, i.e., more accurate experts have higher probability of establishing connections. Equation 3 presents the computation of a node receiving a new connection.

$$\Pi(k) = \frac{1}{\sum_j \phi_j} \cdot \left( \sum_i |\phi_i - \phi_k| \right) \quad (3)$$

Equation 3 is based on error metrics of each node in the network, where $\sum_j \phi_j$ stands for the sum of all nodes errors of the network, $\sum_i |\phi_i - \phi_k|$ is the sum of the absolute deviations between every node in the network when compared to a node $n_k$ and $\phi_k$ is the RMSE metric for this node.

Equation 4 presents the computation of RMSE $\phi_i$ for a given node $n_i$. The variables presented in Equation 4 are: the amount of instances predicted by the inductor $|s|$, where $s$ is a subset of the entire stream $S$; the value obtained by the inductor in the prediction of an instance $i_l$, $o_l$; and the expected value for $i_l$ ($e_l$).

$$\phi_i = \sqrt{\frac{1}{|s|} \sum_{\forall i_l \in s} |o_l - e_l|^2} \quad (4)$$

SFNR assigns weights for experts in the network accordingly to their centrality metrics. The centrality metric $\zeta$ is a user-given parameter. Equation 5 shows the prediction calculation where $\sum_d^N \zeta_d \cdot h_d(i)$ stands for the sum of each expert's prediction for a given instance $i$ ($h_d(x)$) weighted by its centrality measure ($\zeta_d$) and $\sum_k^N \zeta_k$ is the sum of every expert's centrality metrics.

$$H(i) = \frac{\sum_d^N \zeta_d \cdot h_d(i)}{\sum_k^N \zeta_k} \quad (5)$$

Each centrality metric directly affects the accuracy results since each one distributes weights for nodes in different ways. In (Barddal & Enembreck, 2013; Barddal, Gomes, & Enembreck, 2014) authors observed that Eigenvector centrality metric yielded the best average accuracy results for evolving streams. The Eigenvector centrality of an expert $v$ such that $v \in N$ is given by Equation 6, such that if there is a connection between $v$ and $u$, then $A_{vu} = 1$, otherwise $A_{vu} = 0$; $N$ is the set of all experts of the network; $\psi$ is a normalization parameter; and $\rho$ controls how neighbors of $v$ will influence its eigenvector value. Intuitively, the Eigenvector centrality yields higher values for experts that are connected to experts who have higher eigenvector values. In opposition to geodesic-based centrality metrics (such as Betweenness), Eigenvector centrality metric achieved higher accuracy since it does not rank non-central vertices with zero, thus weights are more equally distributed in the network.

$$c_v(\rho) = \sum_{u \in N} (\psi + \rho c_u) A_{vu} \qquad (6)$$

The main hypothesis of SFNR is that in order to obtain a set of experts that maintain low error rates, we must enhance the weight association methods, without waiving diversity of experts. The adaptation of the ensemble into a social network is due the possibility of using graph theory, statistics and algebraic models aiming on providing theoretical robusticity to results, as also providing relationships between ensemble individuals.

We emphasize that SFNR is not bounded to a single base expert. Thus, it is possible to use any online regressor as a base learner. SFNR was developed under MOA framework (Bifet, Holmes, Kirkby, & Pfahringer, 2010).

Algorithm 1 presents the pseudo code for the SFNR algorithm. SFNR expects as input a data stream $S$, which makes available an instance $i$ after each $t$ moments of time, a maximum period error $\theta$, a centrality metric $\zeta$ and the period size $p$.

Firstly, SFNR initializes the network with a single expert $e$. While the stream $S$ is not over, SFNR will split these instances in evaluation periods of size $p$. For every period, instances are obtained from $S$ and predictions are calculated with the aid of a weighted mean based on each expert's prediction. The weights are computed based on a user-given parameter $\zeta$, which determines which centrality metric should be used for polling. SFNR is able to use any centrality metric, but so far the implementation is capable of using: Degree, Betweeness, Closeness, Eigenvector and PageRank.

During a period, all instances are kept in a set of instances, namely $V$. This set $V$ is later used for training a new expert

**Algorithm 1.** SFNR pseudo code

**Input:** A data stream $S$ which makes available an instance $i$ every $t$ moments, a period size $p$, a maximum error threshold $\theta$, a user given error metric $\omega$ and a centrality metric $\zeta$.

**Local Variables:** a network of experts $N = \{n_1, n_2, n_3, ..., n_\aleph\}$, where $\aleph$ is the variable ensemble size, an initial expert $e$, an instance $i = (\vec{x}, y)$, an error accumulator error $\phi$ and a set of instances $V$ used for training a new expert.

1:    $N := \{e\}$
2:    **while** $not\_over(S)$
3:      **for** $j := 1$ **to** $p$
4:        $i := next\_instance(S)$
5:        $prediction := getPrediction(N, i, \zeta)$
6:        $updateErrorMetric(\phi, prediction, i.y)$
7:        $V := V \cup \{i\}$
8:        **for each** $n_j$ **in** $N$
9:          $train(n_j, i)$
10:        **end for**
11:      **end for**
12:      **if** $(\phi > \theta)$
13:        $removeInductor(N)$
14:        $rewiring(N)$
15:        $addInductor(N, V)$
16:        $updateCentralityMetric(N, \zeta)$
17:      **end if**
18:      $V := \emptyset$
19:      $\phi := 0$
20:    **end while**

which will be added to the network. In our approach, after the prediction of a given instance, all experts are trained with the same instance. One could argue that since all experts are trained with the same instance, diversity would be diminished. Nevertheless, since experts are added and removed in different moments in time, their concepts usually do not converge (Barddal, Gomes, & Enembreck, 2014). At the end of a period, error metrics are tested and compared to the user-given threshold $\theta$. If the global error metric is above $\theta$, then the worst expert of the network is removed. When an expert is removed, it is possible that the network becomes a disconnected graph. Thus, a rewiring process (Albert & Barabási, 2002) is used in order to let the network in a connected state once again. Basically, the rewiring process establishes connections between the neighbors of the removed node using the adapted preferential attachment law. Since the network is in a connected state, a new expert will be trained with the instances stored in $V$. Therefore this new expert is added to the network with the adapted preferential attachment law presented earlier in this paper (see Equation 3).

Finally, since the network topology has changed, the centrality metric $\zeta$ is updated for use in the next period.

## ADVANCES ON THE SFNR ALGORITHM

As seen in the previous section, SFNR (Barddal & Enembreck, 2013) was presented where concept drift was detected by determining a period size $p$, which determined the amount of instances that would be evaluated before a network update took place and a user-given maximum error threshold $\theta$. Network updates occurred if an error accumulator $\phi$ was greater or equal to $\theta$. Therefore, the major limitations of SFNR is its high dependency of the user-given parameters $p$ and $\theta$. Thus, we can not assume the user is an expert of the stream domain and also, optimal values of $p$ and $\theta$ may vary during time.

In this paper we present an extension to this work, where we detect drifts using the Adaptive Window (ADWIN) algorithm (Bifet & Gavaldà, 2007), thus, neither of earlier presented parameters are needed. ADWIN is a change detector and estimator that solve the problem of tracking the average of a stream of bits or real-valued numbers. ADWIN keeps a variable-length window ($W$) of recently seen data, based on

**Algorithm 2**. ADWIN pseudo code.

**Input:** a data stream of examples $S$ and a confidence level $\delta$.

**Output:** a window of examples $W$.

1: **Initialize the window** $W$
2: **for each** $i$ **in** $S$
3: $\quad W := W \cup \{i\}$
4: **repeat**
5: $\quad$ drop the oldest element from $W$
6: **until** $|\widehat{\mu_{W_0}} - \widehat{\mu_{W_1}}| < \epsilon_{cut}$ **holds for every split of** $W = W_0 \cdot W_1$

the property that the window has the maximal length possible, yet, statistically consistent with the given hypothesis: "there has been no change in the average value inside the window". ADWIN's only user-given parameter is a confidence bound $\delta$, which indicates how confident we want to be in the algorithm's output decisions. Algorithm 2 presents the pseudo code for ADWIN.

The key part in ADWIN is the definition of $\epsilon_{cut}$ and the statistical test used. The value of $\epsilon_{cut}$ is calculated after Equation 7, where $n$ denotes the size of W, $n_0$ and $n_1$ represent the sizes of $W_0$ and $W_1$ respectively, thus $n = n_0 + n_1$. Let $\widehat{\mu_{W_0}}$ and $\widehat{\mu_{W_1}}$ be the averages of the values $W_0$ and $W_1$.

$$\epsilon_{cut} = \sqrt{\frac{n_0 + n_1}{2} \cdot \frac{4n}{\delta}} \qquad (7)$$

Initially SFNR+ADWIN algorithm instantiates a single expert $e$ for learning the stream (Algorithm 3). While no drifts are detected by ADWIN, instances are retrieved from the stream $S$ and the procedure *getPrediction* (Equation 5) calculates a prediction for each instance weighted on a user-given centrality metric $\zeta$.

Instead of weighting experts' votes by its accuracy in early predictions, weights are updated only when drifts occur. Although experts with higher accuracy tend to establish more connections, we observed that these weights are not updated correspondently to their accuracy since new connections are probabilistic.

After each prediction, every expert in the network is trained with the same instance $i$. Although there is no guarantee that their concepts will not converge, the preferential attachment process helps since an expert with bad accuracy may still establish new connections, increasing its centrality metric.

When a drift is detected, the network evolves. The evolution of the network is divided into: expert removal, rewiring process and expert addition. Firstly, the expert removal process determines whether and which expert in the network should be removed. It is verified if the network size $k$ is greater than a user-given maximum network size $k_{max}$. This parameter was added in order to prevent the network from growing indefinitely. When $k > k_{max}$, the expert with higher RMSE metric in the network is removed and the rewiring process is executed. Since both removal and rewiring processes are completed, a new expert is trained and added to the network using the split window $W$. ADWIN determines the split window $W$.

Finally, once both experts' removals and additions are done, it is possible to calculate the centrality metric $\zeta$ for the next period.

## EXPERIMENTS

### Synthetic Data

For synthetic data experiments, we used a modified hyperplane generator where the prediction function must determine the Euclidian distance between each instance to a random hyperplane (Shaker & Hüllermeier, 2012).

**Algorithm 3.** SFNR + ADWIN pseudo code.

**Input:** a data stream of examples $S$, a maximum network size $k_{max}$ and a confidence level $\delta$.

**Local variables:** a network of experts $N = \{n_1, n_2, n_3, ..., n_\aleph\}$, where $\aleph$ is the variable ensemble size, an initial expert $e$, an instance $i = (\vec{x}, y)$ and a window of examples W.

1:    $N \coloneqq \{e\}$
2:    **while** $not\_over(S)$
3:      **while** $no\ drifts\ are\ detected\ by\ ADWIN(W, \delta)$
4:        $i \coloneqq next\_instance(S)$
5:        $prediction \coloneqq getPrediction(N, i, \zeta)$
6:        **for each** $n_j$ **in** $N$
7:          $train(n_j, i)$
8:        **end for**
9:        $W \coloneqq W \cup \{i\}$
10:      **end while**
11:      **if** $(k > k_{max})$
12:        $removeInductor(N)$
13:        $rewiring(N)$
14:      **end if**
15:      $addInductor(N, V, W)$
16:      $updateCentralityMetric(N, \zeta)$
17:    **end while**

Concept drifts are simulated by mixing two streams with different hyperplanes using the sigmoid function (Equation 8) presented in (Bifet, Holmes, Pfahringer, Kirkby, & Gavaldà, 2009) where $s$ stands for the drift window length, $t$ is the amount of instances already evaluated and $t_0$ is the time of the drift.

$$f(t) = \frac{1}{1 + e^{-s(t-t_0)}} \quad (8)$$

Equation 8 has a derivative at time $t_0$ equal to $f'(t_0) = s/4$ and that $\tan \alpha = f'(t_0)$, thus $\tan \alpha = s/4$. Also, $\tan \alpha = 1/W$ and as $s = 4 \tan \alpha$ then $\alpha = 4/W$, where $W$ stands for the length of the drift window and $\alpha$ is the phase angle.

### Real-World Data

In order to compare algorithms in real-world situations we present results for three datasets presented at UCI Machine Learning repository (*http://archive.ics.uci.edu/ml/*).

*Relative Location of CT Slices on Axial Axis Dataset*

This dataset contains 53,500 images from 74 different patients. Each CT slice is described by two histograms in polar space coordinates. The first histogram describes the location of bone structures in the image while the second presents the location of air inclusions inside of the body. Both histograms are concatenated to form the final feature vector.

*Wine Quality Datasets*

These two datasets are related to both red and white variants of the Portuguese "Vinho Verde" wine where the objective is to determine the quality of the wine based on its attributes (Cortez, Cerdeira, Almeida, Matos, & Reis, 1998).

*Stock Market Data*

We developed a stock market data parser under MOA (Bifet, Holmes, Kirkby, & Pfahringer, 2010). This parser connects to *Yahoo!* to gather data from a given stock in a period of time and desired periodicity, i.e. daily, weekly or monthly. *Yahoo!* supplies data in a CSV (comma-separated values) format file with the following attributes: *Date, Open, High, Low, Close, Volume and Adj Close*. *Date* is the date of that instance. *Open* is the initial value of that stock in the instance given period while *High* and *Low* are the highest and lowest values, respectively. *Volume* is the amount of transfers in which that stock was involved. The data obtained from *Yahoo!* for our experiments represents all the instances in a daily periodicity from January 1st 1996 to January 31st of 2014.

**Experimental Protocol**

All experiments were performed in Intel Xeon w3520 2.67GHz x4 with 12 GB of RAM running Windows Server 2013.

We used the Prequential procedure (Gama & Rodrigues, 2009) where every instance is tested then trained only once. Our option to the Prequential procedure is due to its monitoring of the evolution of performance of models over time, even though it may be pessimistic in comparison to the holdout estimative. Nevertheless, authors in (Gama & Rodrigues, 2009) observe that the Prequential error converges to a periodic holdout estimative (Bifet et al., 2010) when estimated over a sliding window. Along these lines, we determined a sliding window of 100,000 instances for all experiments. Experiments where the number of instances is minor than 100,000 were evaluated with a single window. Table 1 presents the parameters for the Rotating Hyperplane Regression experiments.

Based on (Barddal & Enembreck, 2013), the parameters for AddExp are: Weakest First Pruning Method, a decreasing multiplicative constant $\beta = 0.5$, a factor for new expert weight $\gamma = 0.1$, an expert addition threshold $\tau = 0.05$ and a maximum ensemble size $k = 10$. The parameters for IBLStreams are: Prediction strategy of local linear regression, an internal evaluation window $w = 1000$, the adaptation strategy as AdaptK where $k$ is the varying amount of nearest neighbors to be taken in consideration for prediction inside the interval [4; 40] and $1/d$ distance weighting where $d$ is the Euclidean distance. We also present results for the original SFNR using a period size $p = 1000$, a maximum error threshold $\theta = 0.08$, a maximum network size $k_{max} = 10$ and centrality metric $\zeta$ = Eigenvector. Finally, the SFNR+ADWIN parameters are: confidence bound $\delta = 10\%$, a maximum network size $k_{max} = 10$ and centrality metric $\zeta$ = Eigenvector.

*Table 1. Experiments Configurations.*

| Experiment Identifier | Stream Configuration | | |
|---|---|---|---|
| | # of drifts | Length of drift window(s) ($W$) | Time of drift ($t_0$) |
| RHPR-1 | 1 | 1 | 500,000 |
| RHPR-2 | 1 | 1,000 | 500,000 |
| RHPR-3 | 2 | 1 | 333,333 750,000 |
| RHPR-4 | 2 | 1,000 | 333,333 750,000 |

*Table 2. RMSE Metrics for Experiments*

| Dataset | Root Mean Squared Error (RMSE) | | | | | | | | |
|---|---|---|---|---|---|---|---|---|---|
| | FIMT-DD | AddExp (FIMT-DD) | SFNR (FIMT-DD) | SFNR+ADWIN (FIMT-DD) | AMRules | AddExp (AMRules) | SFNR (AMRules) | SFNR+ADWIN (AMRules) | IBLStreams |
| Red Wine | 0.8357 | 0.8305 | **0.8092** | 0.8183 | 0.9778 | 0.9756 | **0.7541** | 0.7666 | 4.2116 |
| White Wine | 0.9266 | 0.9154 | 0.8778 | 0.8871 | 1.0216 | 0.9875 | 0.8254 | **0.8129** | 2.3152 |
| CT Scans | 20.6363 | 20.5981 | 20.1116 | 20.4516 | 17.7981 | 15.9710 | 12.6784 | 13.1470 | **7.6797** |
| RHPR-1 | **0.0107±0.0002** | 0.0115±0.0011 | **0.0101±0.0003** | 0.0102±0.0005 | 0.0821±0.0008 | 0.0821±0.0023 | **0.0337±0.0003** | 0.0339±0.0005 | 0.1116±0.0016 |
| RHPR-2 | **0.0161±0.0003** | 0.0159±0.0002 | **0.0152±0.0002** | 0.0158±0.0002 | 0.0822±0.001 | 0.0947±0.0022 | **0.0339±0.0004** | 0.0342±0.0007 | 0.1117±0.0002 |
| RHPR-3 | 0.0258±0.0002 | 0.0264±0.0002 | **0.0242±0.0001** | 0.0243±0.0001 | 0.1042±0.0003 | 0.1177±0.0003 | **0.0547±0.0004** | 0.0551±0.0003 | 0.1393±0.0002 |
| RHPR-4 | 0.0258±0.0003 | 0.0262±0.0003 | **0.0247±0.0002** | 0.0247±0.0003 | 0.1042±0.0003 | 0.1206±0.0003 | **0.0527±0.0005** | 0.0532±0.0003 | 0.1393±0.0003 |

*Table 3. RMSE Metrics for Stock Market Experiments.*

| Experiment Identifier | Root Mean Squared Error (RMSE) | | |
|---|---|---|---|
| | EMA | SFNR (EMA) | SFNR+ADWIN (EMA) |
| SM-YHOO | 10.1380 | **9.0282** | 9.0282 |
| SM-GOOG | 14.3101 | **10.3351** | 14.0331 |
| SM-MSFT | 3.9050 | 3.7506 | **3.7406** |
| SM-XOM | 2.4174 | 2.2408 | **2.2397** |

Whereas, for the stock market experiments, we only evaluated the Exponential Moving Average Algorithm, with a window size $w = 5$ and the SFNR+ADWIN using the same Moving Average Algorithm, yet, using a confidence level $\delta = 10\%$, a maximum network size $k_{max} = 10$ and centrality metric $\zeta$ = Eigenvector. Other algorithms were not evaluated hence the dataset attributes already bias the objective value since the *Close* attribute is always between *Low* and *High*.

**Results**

Table 2 presents the RMSE metrics obtained in our experiments. Comparing the results of algorithms for the Red Wine and White Wine datasets we can observe that both SFNR and SFNR+ADWIN improved both FIMT-DD and AMRules RMSE metrics, yet, at the CT Scans experiment, IBLStreams outperformed all others with substantial difference. The mean values for the RHPR experiments were calculated based on 50 executions varying the pseudo-randomization seed.

With the aid of Lielliefors (Liellierfors, 1967) normality test we determined that the obtained RMSE distributions do respect normality. Therefore, we compared SFNR+ADWIN with the other algorithms using conventional paired testing (confidence of 95%). This comparison was made in two steps. Firstly, we compared the FIMT-DD, IBLStreams and the ensemble methods using FIMT-DD as a base learner (AddExp, SFNR and

## Figure 3. RHPR-1 RMSE metric during the stream.

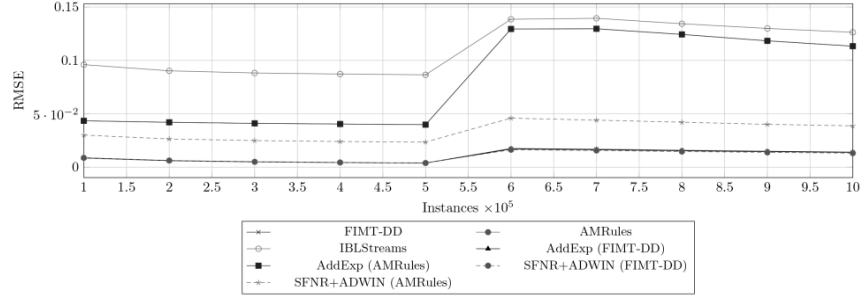

## Figure 4. RHPR-2 RMSE metric during the stream.

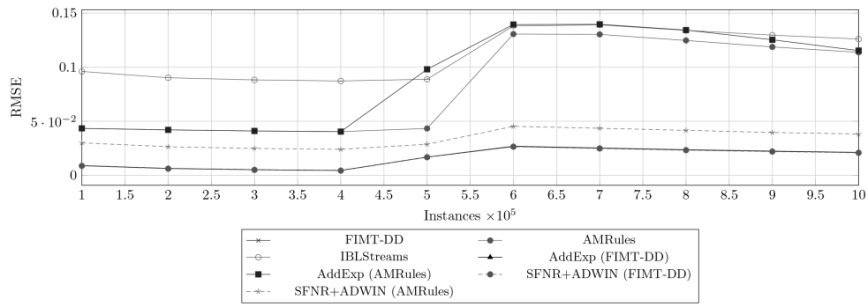

SNFR+ADWIN). Later, we compared AMRules, IBLStreams and the ensemble methods using AMRules as a base learner.

Table 2 also presents the results of our comparison where bolded face indicates statistical better results in comparison to other algorithms. Where no conclusive statistical difference was found, we bold-face all similar algorithms.

First, we compared SFNR+ADWIN (FIMT-DD) to a single FIMT-DD, AddExp (FIMT-DD) and IBLStreams, where we concluded that there is no statistical differences between them. Then, in another comparison between SFN, SFNR+ADWIN (AMRules), AddExp (AMRules) and AMRules one could see that both SFNR and SFNR+ADWIN boosted AMRules performance showing relevant statistical differences, especially in experiments with two concept drifts.

In the stock market experiments, presented in Table 3, we can observe that both SFNR and SFNR+ADWIN were able to improve the base Exponential Moving Average algorithm accuracy in all cases, where SFNR and SFNR+ADWIN presented equal RMSE metrics for SM-YHOO experiment, SFNR had better result for the SM-GOOG experiment and SFNR-ADWIN showed better results for both SM-MSFT and SM-XOM experiments.

In Table 4 we present the average processing time for the experiments. One could see that SFNR+ADWIN present competitive processing time when compared to AddExp and the original SFNR in all configurations. Thus, one can see that the usage of ADWIN as a drift detector does not jeopardize the method in terms of processing time. We also emphasize the SFNR+ADWIN associated with AMRules, where the ensemble outperformed a single expert both in RMSE and processing time. Since AMRules has a rule set that tends to grow, this shows that an ensemble of AMRules where experts are added and removed tend to replace experts with larger rules sets, thus diminishing processing time for triggering prediction rules.

Figure 5. RHPR-3 RMSE metric during the stream.

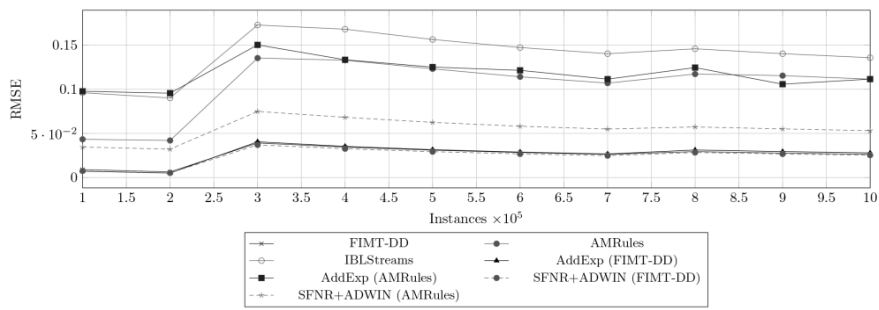

Figure 6. RHPR-4 RMSE metric during the stream.

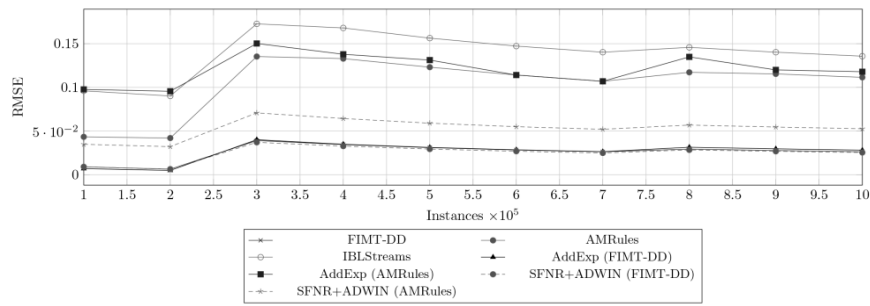

*Table 4. Average Processing Time in Seconds.*

| Dataset | Average Processing Time ($s$) | | | | | | | | |
|---|---|---|---|---|---|---|---|---|---|
| | FIMT-DD | AddExp (FIMT-DD) | SFNR (FIMT-DD) | SFNR+ADWIN (FIMT-DD) | AMRules | AddExp (AMRules) | SFNR (AMRules) | SFNR+ADWIN (AMRules) | IBLStreams |
| Red Wine | **0.2378** | 0.2862 | 0.2994 | 0.2943 | 0.3791 | 0.3987 | 0.2600 | **0.2593** | 0.7401 |
| White Wine | **0.4136** | 0.7678 | 0.5921 | 0.5879 | 1.2185 | 1.4578 | 0.5228 | **0.5211** | 3.8297 |
| CT Scans | **17.4554** | 52.3670 | 34.7577 | 34.7541 | 25.3513 | 50.2039 | 24.7102 | **24.7097** | 385.9726 |
| RHPR-1 | **22.8104** | 72.5467 | 59.0566 | 59.0516 | 281.4294 | 440.7845 | 204.7866 | **204.7831** | 6381.8066 |
| RHPR-2 | **22.6675** | 72.1395 | 59.1576 | 59.1540 | 277.7916 | 441.2579 | 204.8207 | **204.8191** | 6297.9164 |
| RHPR-3 | **23.0172** | 72.9476 | 60.0165 | 60.0159 | 284.4652 | 445.0081 | 206.7701 | **206.7694** | 6394.9312 |
| RHPR-4 | **23.0346** | 73.0056 | 60.0478 | 60.0450 | 284.7894 | 445.0089 | 206.1392 | **206.1375** | 6395.0164 |

Besides, another conclusive result obtained is that the original SFNR in comparison with SFNR+ADWIN are statistically equal, thus, SFNR+ADWIN is a better choice due to its diminished amount of parameters.

Apart from average error rates, it is important to observe the RMSE evolution during the entire stream, therefore examining the algorithms behavior before, during and after drift happens, observing the adaptability of the methods to the drifts. Figures 3 through 6 present the evolution of the stream Prequential errors for the RHPR experiments. In these Figures we can see that FIMT-DD, AddExp (FIMT-DD) and SFNR+ADWIN (FIMT-DD) show similar behaviors, enlightening the inherent adaptability of the FIMT-DD algorithm to concept drifts. Yet, when comparing AMRules, AddExp (AMRules) and SFNR+ADWIN (AMRules), one could see that SFNR+ADWIN learner improves the RMSE metrics during the whole streams, especially after the concept drift, where these metrics barely changes. In order to preserve readability, Figures 3 through 6 ommit results for the original SFNR, since it is similar to SFNR+ADWIN.

## CONCLUSIONS AND FUTURE WORK

SFNR+ADWIN results present accuracy improvements in comparison with the state-of-the-art algorithms in a variety of experiments, especially in concept drift situations. Since the majority of algorithms for handling concept drift aim on the classification task, we believe our proposal is feasible and demonstrates the power of social ensemble methods in regression tasks. Jutting the synthetic data streams, the usage of a real stream, i.e. stock market data, CT slices and Wine Quality; allowed us to recognize the effectiveness SNFR+ADWIN in real-world situations.

In future works we plan on widening our research in time-series prediction, multi-label stream classification and data stream clustering. A detailed study of SFNR+ADWIN in different stream configurations with concept drift will be developed in order to evidence the semantics between each centrality metric and the ensemble topology and votes. We also plan to study the experts' average lifetime, other topology metrics and concept drift detection methods in order to improve SFNR+ADWIN.